\documentclass[conference]{IEEEtran}
\IEEEoverridecommandlockouts

\usepackage{caption}
\usepackage{cite}
\usepackage{amsmath,amssymb,amsfonts}
\usepackage{algorithmic}
\usepackage{graphicx}
\usepackage{textcomp}
\usepackage{xcolor}

\usepackage{hyperref}
\usepackage{url}
\usepackage{subcaption} 

\usepackage{graphicx}
\usepackage{booktabs}
\usepackage{bm}
\usepackage{makecell}
\usepackage{multirow}

\usepackage{stackengine}
\newsavebox\mybox
\newcommand\Includegraphics[2][]{\sbox{\mybox}{%
  \includegraphics[#1]{#2}}\abovebaseline[-.5\ht\mybox]{%
  \addstackgap{\usebox{\mybox}}}}

\usepackage{pifont}
\newcommand{\cmark}{\ding{51}}%
\newcommand{\xmark}{\ding{55}}%

\begin{document}

\title{Disentangled representations of microscopy images}

\author{%
  Jacopo Dapueto \quad Vito Paolo Pastore \quad Nicoletta Noceti \quad Francesca Odone \\
  \texttt{jacopo.dapueto@edu.unige.it} \\
  \texttt{\{vito.paolo.pastore,nicoletta.noceti,francesca.odone\}@unige.it}\\
  MaLGa-DIBRIS, Università degli studi di Genova, Genova, Italy \\
}

\maketitle

\begin{abstract}
Microscopy image analysis is fundamental for different applications, from diagnosis to synthetic engineering and environmental monitoring.  
Modern acquisition systems have granted the possibility to acquire an escalating amount of images, requiring a consequent development of a large collection of deep learning-based automatic image analysis methods.
Although deep neural networks have demonstrated great performance in this field, interpretability — an essential requirement for microscopy image analysis — remains an open challenge.  

This work proposes a Disentangled Representation Learning (DRL) methodology to enhance model interpretability for microscopy image classification. 
Exploiting benchmark datasets from three different microscopic image domains (plankton, yeast vacuoles, and human cells), we show how a DRL framework, based on transferring a representation learnt from synthetic data, can provide a good trade-off between accuracy and interpretability in this domain. 
\end{abstract}

\begin{IEEEkeywords}
 microscopy images, disentangled representations, transfer learning, interpretability.
\end{IEEEkeywords}

\section{Introduction}
\label{sec:intro}

The analysis of microscopy images is crucial for biomedical research \cite{liu2021survey}, from histopathological diagnosis to analysis of cell organelles and classification of microorganisms. Manual analysis has become impractical in recent years, given the large volumes of images acquired through advanced acquisition systems \cite{pastore2023efficient}. Consequently, deep learning has been extensively applied in the microscopy domain for different image analysis tasks \cite{xing2017deep}, including classification \cite{liu2022aimic,firat2024classification}, segmentation \cite{greenwald2022whole,mckinley2022miriam}, and object detection \cite{rivas2020automatic, hung2020keras,kumar2023advances}. On the one hand, deep neural networks (DNNs) have achieved great performance for microscopy image-related tasks, generally outperforming conventional approaches based on handcrafted features \cite{plissiti2018sipakmed, pastore2023unsupervised}. On the other hand, given the intrinsic DNN's \textit{black box} nature, predictions lack interpretable human-reliable insights, highly desirable in this specific domain of application \cite{krishna2023interpretable, tavolara2023one}. Consequently, providing interpretable deep learning methods for biological images is a pressing research challenge \cite{bera2019artificial}. 

In this work, we propose a Disentangled Representation Learning (DRL) framework \cite{bengio2013representation, locatello2019challenging, higgins2017beta, wang2023disentangled} as a potential method to enhance DNN's interpretability in this context.      
DRL aims to learn models that can identify and disentangle underlying Factors of Variation (FoVs) hidden in the observable data,  encoding them in an \textit{interpretable}  and compact way, partially independently from the task at hand. This allows to enhance robustness, and generalization capacity across various tasks \cite{kulkarni2015deep, chen2016infogan, bengio2013representation, zhu2021and, locatello2019fairness, van2019disentangled, wang2023disentangled}. 
DRL was studied in depth in \cite{chen2016infogan} and \cite{higgins2017beta}, and following these works many attempts have been made to learn effectively disentangled representations. \cite{kahana2022contrastive} introduces a contrastive learning paradigm, whereas \cite{lin2020infogan} proposes a contrastive regularization approach for disentangled GANs. The authors in \cite{song2024flow} introduce Flow Factorized Representation Learning which defines a set of latent flow paths that correspond to sequences of different input transformations that resemble the FoVs, while
\cite{renlearning} leverages pretrained generative models for discovering traversal directions as factors with contrastive learning. The authors in \cite{yang2024disdiff} disentangle the gradient fields of the Diffusion Probabilistic Models to discover factors automatically, but finding that the representation may not be easily interpretable by humans.

While the earliest approaches face disentanglement in an unsupervised fashion, without any explicit definition of the FoVs, it has been shown that weakly-supervised approaches to disentanglement provide superior results, see for instance Ada-GVAE  \cite{locatello2020weakly,pmlr-v139-fumero21a}. However, they find limited applicability due to the general lack of knowledge on FoVs characterizing real data. Transfer learning offers a suitable solution, by transferring a disentangled representation from a Source dataset -- where FoVs are known -- to a Target one -- where FoVs might be unknown. {In this direction, a recent study }\cite{dapueto2024transferring}{ shows that a disentangled representation learnt from a synthetic dataset can be transferred to a real one preserving a partial level of disentanglement. Although promising, the analysis is limited to real datasets whose FoVs are controlled and known a priori.}

{ In this work, we move a step further in assessing the applicability of weakly supervised DRL to real-world tasks, by addressing} the specific problem of enhancing interpretability of single cell microscopy image classification, {where the FoVs are only partially known.} { On the methodology side, lacking in the literature examples of DRL in real scenarios, this domain allows us to deal with a real but controlled task, where the number of possible FoVs is limited compared with conventional pictorial images. On the application side, we propose an automatic procedure for learning interpretable representations that could be beneficial in a field where interpretability is as important as accuracy.} 

\begin{figure}[tb]
    \centering

    \begin{subfigure}[t]{0.48\linewidth}
    \centering
    \includegraphics[width=0.18\linewidth]{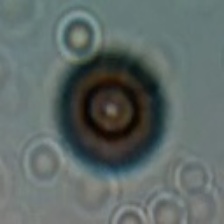}\hfill
    \includegraphics[width=0.18\linewidth]{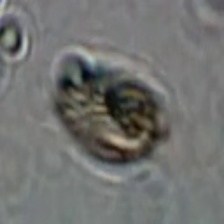}\hfill
    \includegraphics[width=0.18\linewidth]{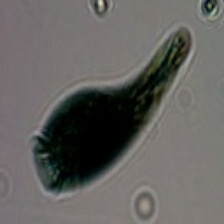}\hfill
    \includegraphics[width=0.18\linewidth]{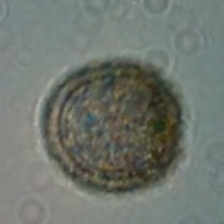}\hfill
    \includegraphics[width=0.18\linewidth]{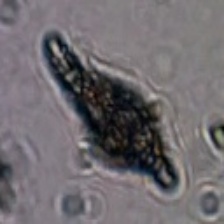}
   \caption{Lensless}
   \label{fig:lensless_samples}
    \end{subfigure}  
    \hfill 
    \begin{subfigure}[t]{0.48\linewidth}
    \centering
    \includegraphics[width=0.18\linewidth]{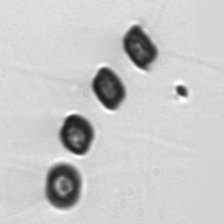}\hfill
    \includegraphics[width=0.18\linewidth]{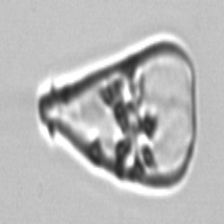}\hfill
    \includegraphics[width=0.18\linewidth]{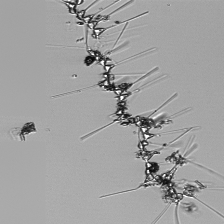}\hfill
    \includegraphics[width=0.18\linewidth]{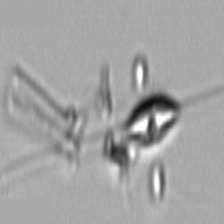}\hfill
    \includegraphics[width=0.18\linewidth]{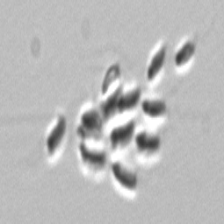}
   \caption{ WHOI15}
   \label{fig:whoi15_samples}
    \end{subfigure}  
        \medskip
    \begin{subfigure}[t]{0.48\linewidth}
    \centering
    \includegraphics[width=0.18\linewidth]{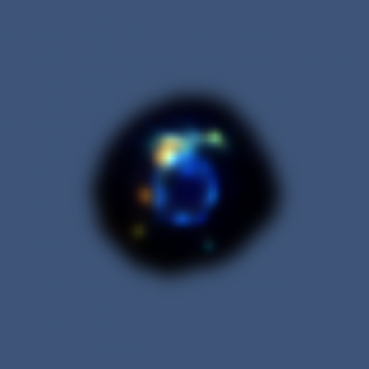}\hfill
    \includegraphics[width=0.18\linewidth]{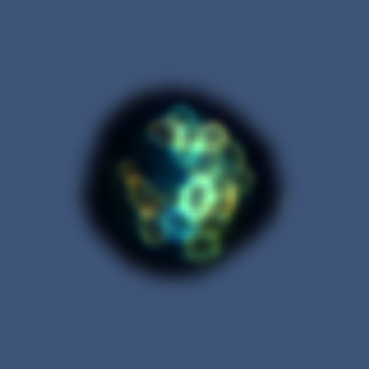}\hfill
    \includegraphics[width=0.18\linewidth]{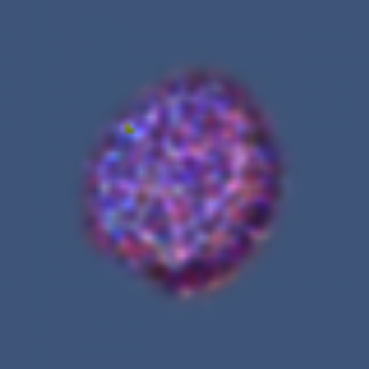}\hfill
    \includegraphics[width=0.18\linewidth]{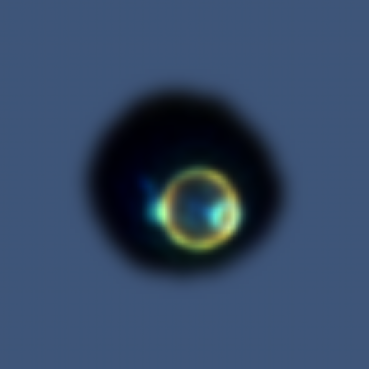}\hfill
    \includegraphics[width=0.18\linewidth]{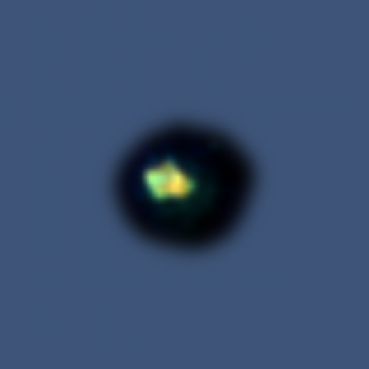}
   \caption{ Vacuoles}
   \label{fig:vacuoles_samples}
    \end{subfigure}  
    \hfill
    \begin{subfigure}[t]{0.48\linewidth}
    \centering
    \includegraphics[width=0.18\linewidth]{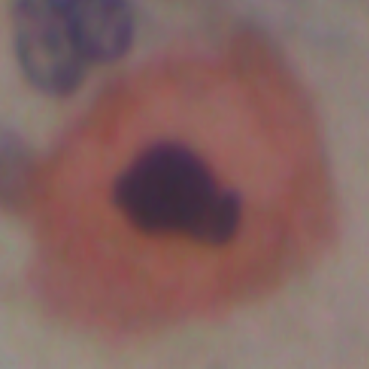}\hfill
    \includegraphics[width=0.18\linewidth]{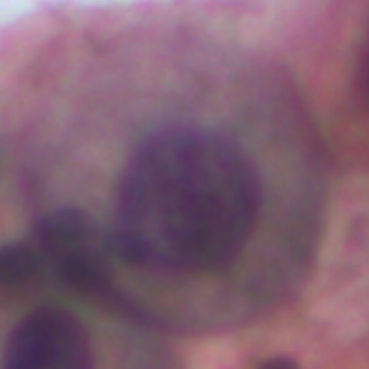}\hfill
    \includegraphics[width=0.18\linewidth]{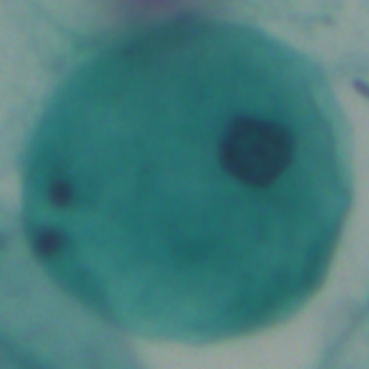}\hfill
    \includegraphics[width=0.18\linewidth]{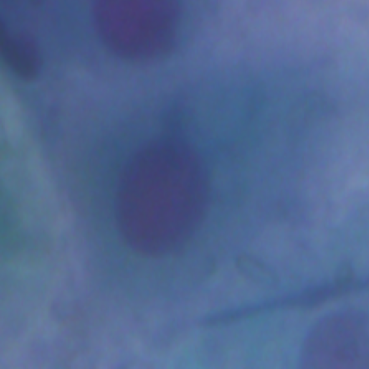}\hfill
    \includegraphics[width=0.18\linewidth]{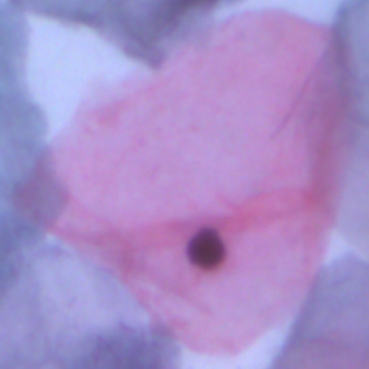}
   \caption{Sipakmed}
   \label{fig:sipakmed_samples}
    \end{subfigure}  
    \caption{5 random samples for each Target dataset}
    \label{fig:datasets}
\end{figure}

Specifically, we adopt datasets coming from three different biological domains: plankton microorganisms, budding yeast vacuoles, and human cancer cells (see Fig. \ref{fig:datasets}). They differ in acquisition systems and type of imaged cell but can be semantically described in terms of relatively simple morphological factors (e.g. texture, shape, color, scale, and other morphological features), thus offering a perfect benchmark for our scope. 
At the same time,  microscopy datasets \cite{pastore2020annotation,ciranni2024anomaly,sosik2015annotated}, like most real-world data,  are not associated with any specific FoVs annotation. 
To overcome this problem, we take inspiration from \cite{dapueto2024transferring} and transfer a latent representation from an Ada-GVAE trained on a different Source dataset to a Target microscopy dataset. 
As a source, we build a simple synthetic dataset, Texture-dSprite, whose (annotated) FoVs may be appropriate to represent the morphological factors we are interested in.
As for the interpretation of the results we exploit the availability in the literature of several handcrafted features computed on the same datasets, that we use as a reference, computing the correlation between these features and the learned dimensions. { In this sense, our work aims to learn the FoVs of the microscopy data following a fully data-driven approach instead.  
}

Moreover, inspired by recent works on unsupervised learning from biological image datasets \cite{pastore2023efficient}, instead of using the images directly as inputs of Ada-GVAE, we provide a projection into a large-dimensional vector of deep features, obtained by a ViT16b model pretrained with DINO self-supervised approach on ImageNet.
With our analysis, we demonstrate that our approach achieves not only good classification accuracy but also disentanglement performances comparable to those learned with synthetic datasets, thus enhancing the interpretability of the learned representations. 

To summarize, the main contributions of this paper are the following:
\begin{itemize}
{
\item We assess a weakly-supervised Disentangled Representation Learning procedure 
on real datasets, with unknown or partially known FoVs. We focus on microscopy images, where disentanglement is a desired property, the data are complex, but the FoVs are somewhat controlled in number. {We observe that by means of a transfer learning paradigm, we can achieve good disentanglement performances on real datasets without FoVs annotation.}
\item { We adopt an input representation based on pretrained deep features in the DRL framework.} Our results on microscopy image datasets show that such design choice allows us to significantly increase the classification accuracy of our interpretable DRL framework with respect to the usage of plain images.
}
\end{itemize}
To the best of our knowledge, this work represents the first application of DRL to real-world datasets and the first attempt of learning disentangled representations from pretrained features.

We believe this paper could serve as a foundation for integrating interpretability { derived from disentanglement }into deep learning frameworks for microscopy image classification tasks, and more broadly, for real-world applications.
The remainder of the paper is organized as follows: in Section \ref{method} we describe the proposed DRL methodology, in Section \ref{experiments} we describe the microscopy image datasets used in this work, provide implementation details, and report the obtained results, evaluating interpretable insights for each dataset, finally discussed in Section \ref{conclusion}.

\section{The proposed medology}\label{method}

\begin{figure}[tb]
\centering
\includegraphics[width=1.0\linewidth]{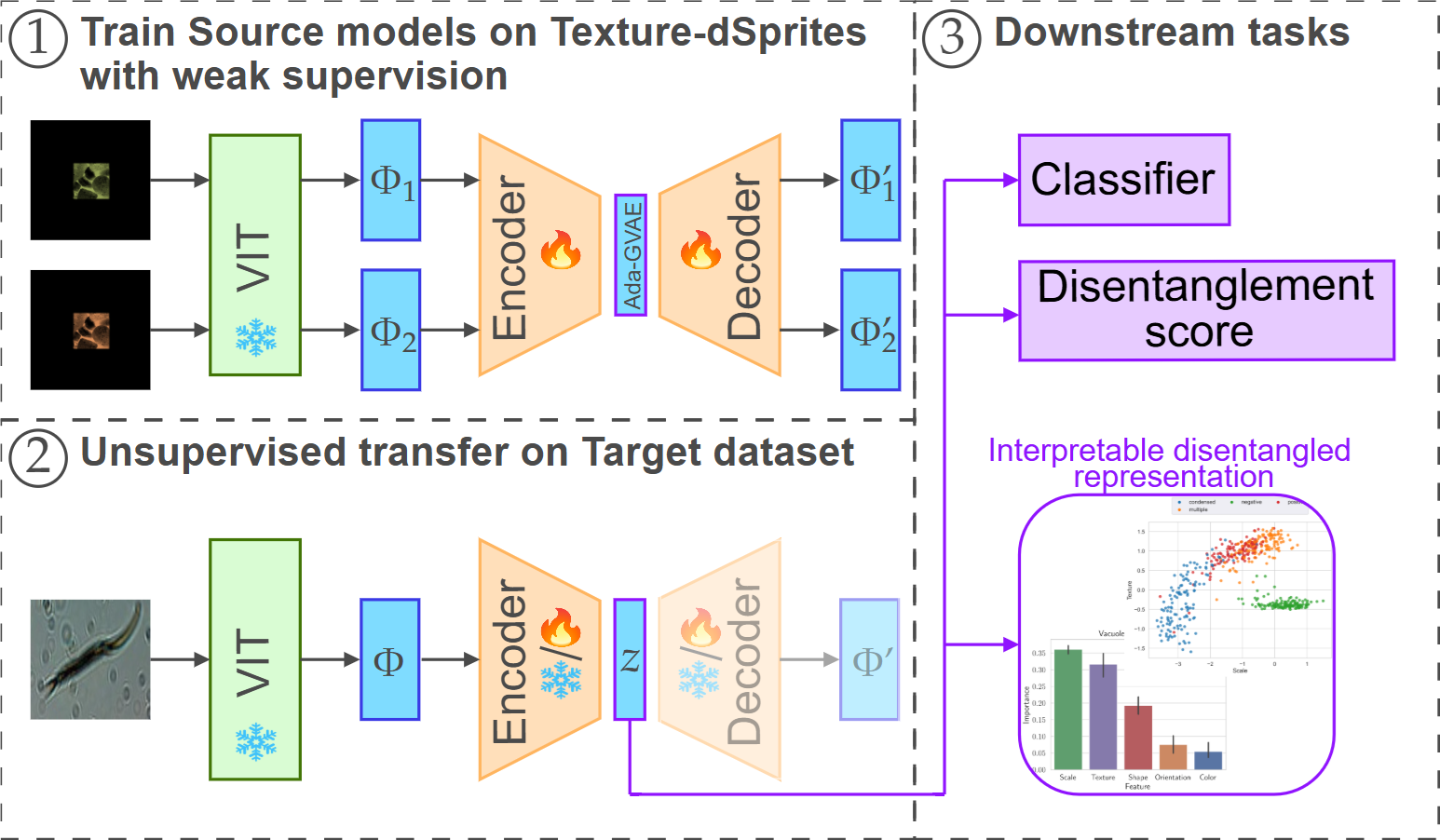}
\caption{ 
A visual sketch of our methodology. We  project each image into a high-dimensional deep feature vector using a pretrained network (ViT16b model pretrained with DINO self-supervised approach on ImageNet1K). Our approach includes 3 main steps. (1) We learn a disentangled model with weak-supervision using Ada-GVAE, using a Source annotated dataset (Texture-dSprite in this case). (2) Then, we transfer the pretrained disentangled model to a real Target dataset (microscopy images in this work). (3) We evaluate the quality of the disentangled model and of the corresponding representation using disentanglement scores and the classification accuracy in a downstream task (the one associated with the Target dataset).
}
\label{fig:pipeline}
\end{figure}

Although the majority of existing methods may be based on different definitions of disentanglement (see for instance \cite{bengio2013representation,higgins2018towards,suter2019robustly}) there is a general agreement that, not surprisingly, disentanglement can be better achieved with some level of supervision on the FoVs \cite{locatello2019challenging}. However, in our target scenario, i.e. microscopy real data, there is no availability of benchmark datasets with this type of annotation. \\
Therefore, in this work we exploit a transfer learning paradigm, to transfer a disentangled model -- trained with weak supervision on a Source dataset where annotation of FoVs is available -- to a Target real dataset -- where we may assume FoVs to be present in the data but they are unknown. To this purpose, we follow the methodology proposed in \cite{dapueto2024transferring}, according to the pipeline in Fig. \ref{fig:pipeline}. 

\subsection{Transferring disentangled representations to real data}

In contrast to the large majority of previous works, instead of disentangling the features extracted from the raw images, we first project the images into a large vector of deep features $\Phi$, by using a pretrained network. Although different choices are possible, we empirically observed that in our scenarios the ViT16b model pretrained with DINO self-supervised approach on ImageNet1K \cite{Caron_2021_ICCV} is the most appropriate since it better captures the complexity of the Source data (see the comparison in Table \ref{tab:backbones}). This choice has been inspired by recent works on microscopy image analysis, showing that such models provide rich and discriminative features for the task at hand,  \cite{pastore2023efficient, kyathanahally2022ensembles}, generally outperforming models trained from scratch \cite{renlearning}. \\
{ Following the same procedure as in \cite{dapueto2024transferring},} we derive the latent disentangled representation using Ada-GVAE \cite{locatello2020weakly} on the Source dataset with annotated FoVs. More specifically, the input of the VAE is a pair 
$\Phi(\pmb{x}_1)$ and $\Phi(\pmb{x}_2)$ for which the images $\pmb{x}_1$ and $\pmb{x}_2$ are sampled from the dataset so that they differ of \textit{k} FoVs, with \textit{k} fixed.\\
Then, we transfer the representation on the Target datasets by finetuning the  models 
on the real unsupervised data with $\beta$-VAE \cite{higgins2017beta}.{  Although}{ we adopted Ada-GVAE and $\beta$-VAE { as in \cite{dapueto2024transferring}, the main difference of our approach is in the choice of the input:} we consider deep features $\Phi$ produced by DINO instead of the RGB images adopted in the previous approach. }

\begin{table}[tb]
\caption{Texture-dSprites: the FoV generating the dataset (Left), examples randomly generated from the FoVs (Right)}

\resizebox{\linewidth}{!}{
\begin{tabular}{cc}
\Large \textbf{Dataset FoVs}& \Large \textbf{Random Samples} \\
\begin{tabular}{ll}
\toprule
\multicolumn{2}{c}{Texture-dSprites} \\
\midrule
\textbf{FoV} & \(\bm{\#}\) \textbf{values} \\
\cmidrule{1-2}
Texture & 5 \\
Color &  7    \\
Shape &  3    \\
 Scale &  6    \\
 Orientation & 40 \\
PosX & 32\\
PosY & 32 \\    
\bottomrule
\end{tabular}& 
 \Includegraphics[width=0.40\textwidth]{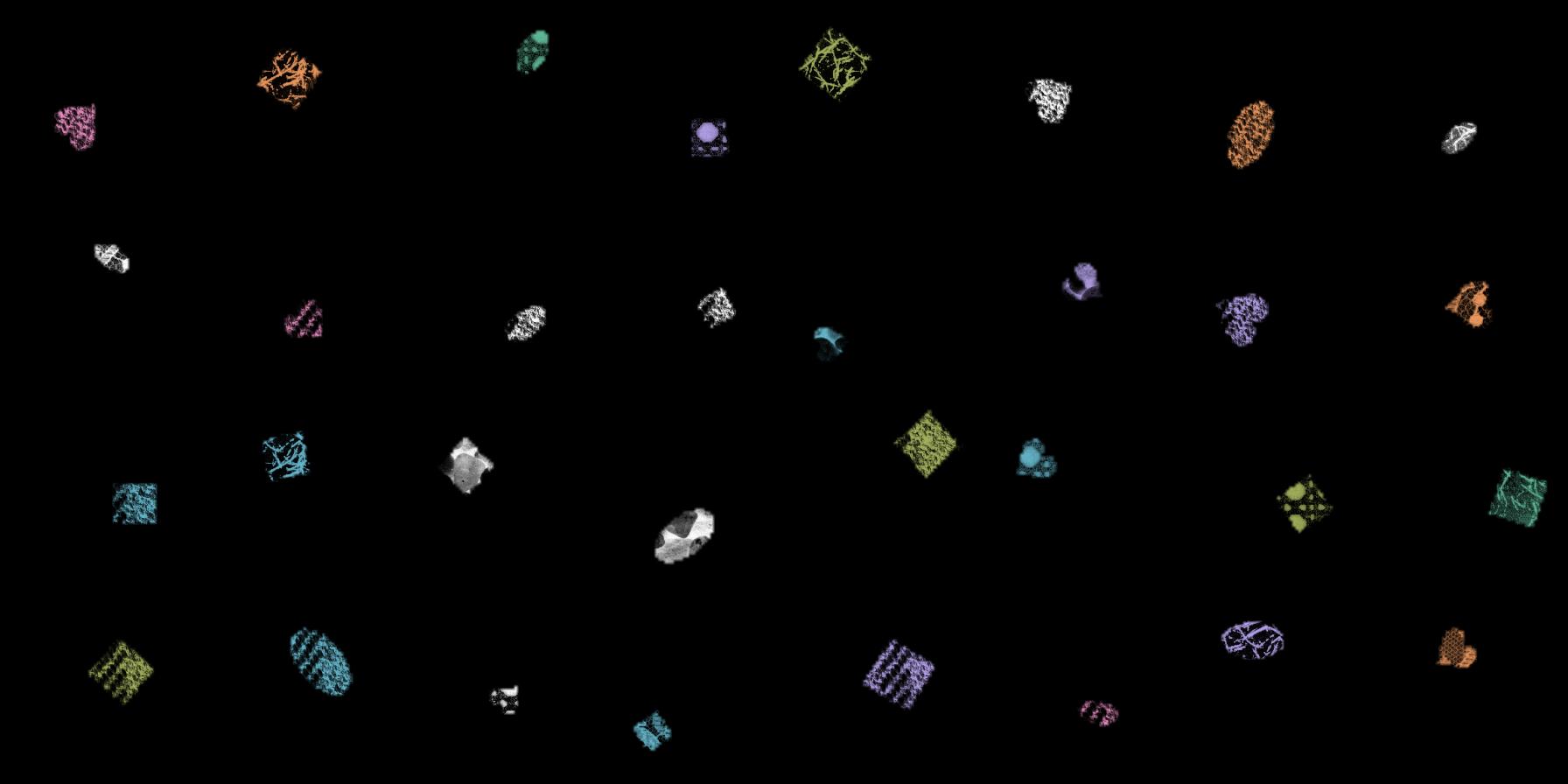} \\

\end{tabular}
}

\label{fig:dsprites}

\end{table}

\subsection{Disentanglement evaluation methods}

For a quantitative evaluation of  disentanglement, there is common agreement on the fact that a disentangled representation should satisfy the following properties
\cite{DBLP:conf/iclr/Do020,van2019disentangled,bengio2013representation}.
 \textbf{Modularity}: a factor influences only a portion of the representation space, and only this factor influences this subspace \cite{ridgeway2018learning,eastwood2018framework}.
   \textbf{Compactness} or completeness: the subset of the representation space affected by a FoV should be as small as possible (ideally, only one dimension) \cite{ridgeway2018learning}. 
   \textbf{Explicitness}: DR should explicitly describe the factors, thus it should allow for an effective FoVs classification \cite{DBLP:journals/corr/Ridgeway16}.

 We evaluate the quality of  disentanglement in our transfer learning scenario by analysing the disentanglement score and the accuracy of a downstream classification task, whose formulation depends on the specific real Target dataset (see Fig. \ref{fig:pipeline}, right).
{ Among the several disentanglement scores from the literature we consider a selection that allows us to capture the different properties mentioned above. More specifically, our analysis includes  }DCI, measuring Modularity \cite{eastwood2018framework}, MIG, evaluating  Compactness \cite{chen2018isolating}, and OMES  incorporating both Modularity and Compactness \cite{dapueto2024transferring}. The latter also facilitates the interpretability of the results, and for this reason well suits our needs.
The downstream classification task  
allows us to evaluate the descriptive power of the representation (Explicitness).  To this purpose, we rely on a simple classifier that does {not} incorporate any further representation learning step.

\begin{figure}[tb]
\centering

    \begin{subfigure}[t]{\linewidth}
    \includegraphics[width=\linewidth]{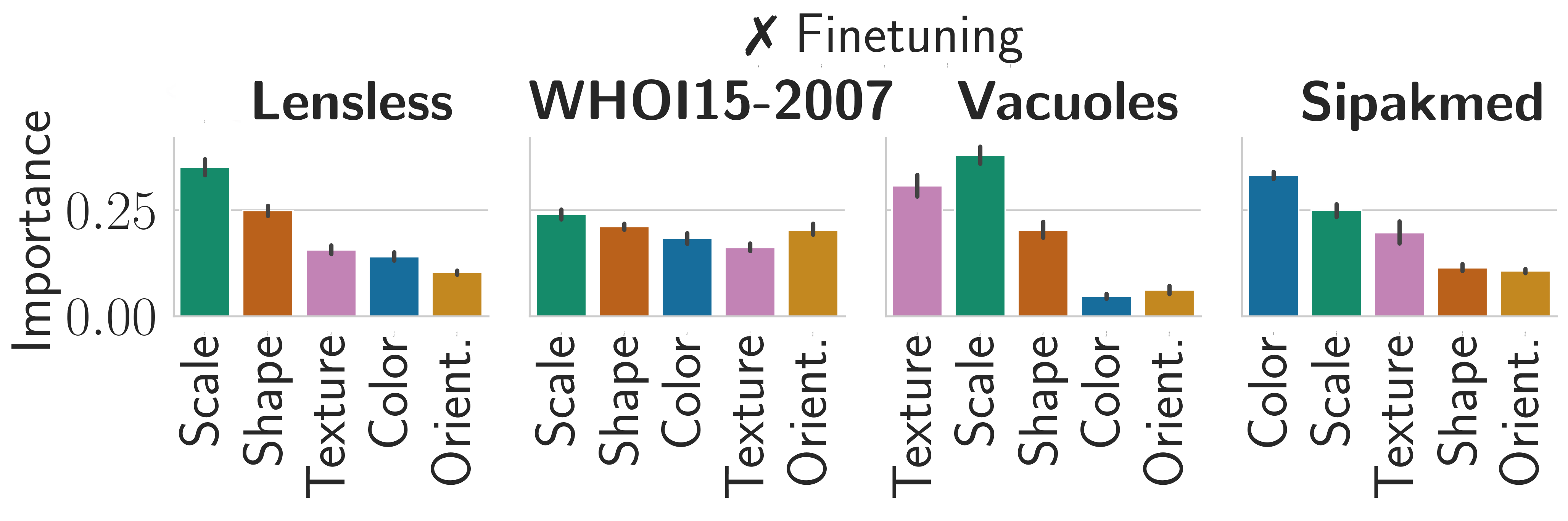}
    \caption{}
    \label{fig:importance_NFT}
    \end{subfigure}
    
    \begin{subfigure}[t]{\linewidth}
    \includegraphics[width=\linewidth]{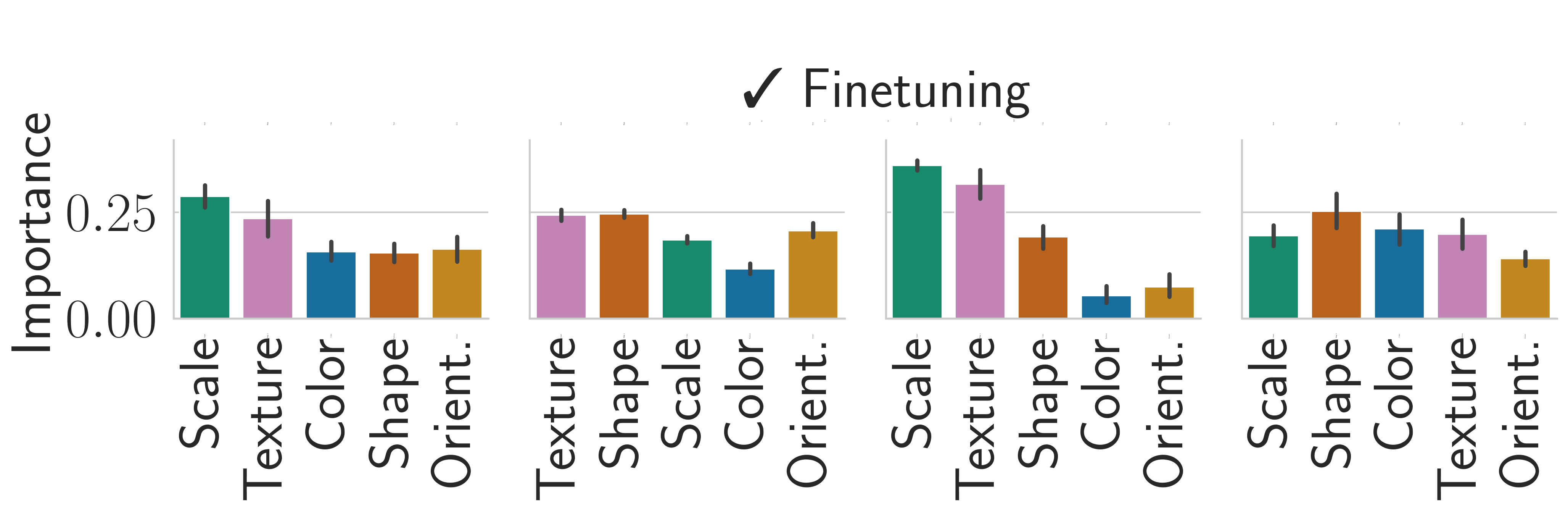}
    \caption{}
    \label{fig:importance_FT}
    \end{subfigure}

\caption{The mean and SD of the feature importance without (Fig. \ref{fig:importance_NFT}) and with (Fig. \ref{fig:importance_FT}) finetuning. These barplots refer to the GBT models trained from $\Phi$. }
\label{fig:lensless_feature_importance}

\end{figure}

\section{Experimental analysis}\label{experiments}

\subsection{Datasets}

\subsubsection*{Real Target datasets\footnote{We will provide the balanced train-test splits (80\%, 20\%) for datasets without native test sets.}}

\textbf{Plankton Lensless} microscope dataset \cite{pastore2020annotation} (Fig. \ref{fig:lensless_samples}) consists of images acquired using a
lensless microscope, extracted from 1-minute videos. It includes 10 classes, with 640 color images each. The dataset includes precise binary masks of each sample so that it is possible to suppress the background and compute handcrafted features like scale, shape, and mean color.\\ 
\textbf{Plankton WHOI15} \cite{ciranni2024anomaly} (Fig. \ref{fig:whoi15_samples}) is a subset of the WHOI dataset \cite{sosik2015annotated}, including 15 classes acquired among 4 years of acquisition (2007-2010). In our experiments, we select the year 2007 subset. With respect to Lensless, this dataset is more challenging due to its fine granularity with high intraclass variability. All the images are grayscale with varying sizes. Segmentation masks, handcrafted features, and train-test splits are not available. 
\textbf{Budding yeast vacuoles}  \cite{pastore2023unsupervised} (Fig. \ref{fig:vacuoles_samples}) dataset includes a total of 998 fluorescence budding yeast vacuole images, extracted from acquired 3D stacks, already divided into training (775) and test (205) sets. Each 3D data volume is converted into a 2D projection in which depth is encoded by color. The dataset is labelled into 4 different morphotypes: single vacuole, multiple, condensed, and negative (or dead cells). A set of handcrafted shape- and texture-based features is available for this dataset (see \cite{pastore2023unsupervised})\\
\textbf{Sipakmed Human Cells} dataset \cite{plissiti2018sipakmed} (Fig. \ref{fig:sipakmed_samples}) consists of 4049 images of isolated cells that have been manually cropped from 966 cluster cell images of Pap smear slides. These images were acquired through a camera adapted to an optical microscope. The cell images are divided into five categories including normal, abnormal and benign cells. A set of handcrafted features describing the nucleus and cytoplasm is available for this dataset (see \cite{plissiti2018sipakmed}).\\
\subsubsection*{Source dataset}

{Even if complete knowledge or annotation is not available for the FoVs in the Target dataset, its general properties may give useful insights into the possible semantics of the FoVs, facilitating the transfer of a disentangled model from a Source dataset.
Having this in mind,} as a Source dataset, we adopted a dataset we generated for the purpose, which we call
{\bf Texture-dSprites}.  It is an annotated synthetic dataset, built as an extension of dSprite \cite{dsprites17} by incorporating 5 different \texttt{Textures} from \cite{abdelmounaime2013new}, in addition to the original FoVs (3 \texttt{Shapes},  7  \texttt{Colors}, 6 values of \texttt{Scale} and 40 values for the \texttt{Orientation}). The original dSprites also include 32 x-positions and 32 y-positions. However, since in our target datasets the object of interest is always centered these factors can be neglected. Table \ref{fig:dsprites} reports the dataset FoVs and some random samples.

\begin{table}[tb]
\caption{Accuracy ($\%$) and SD{ of the classifiers trained on the disentangled representation extracted from the VAE}. 
}
\centering
\resizebox{\linewidth}{!}{
\begin{tabular}[t]{cccccc}
\toprule
 & \multicolumn{2}{c}{\textbf{\textcolor{red}{\xmark}
 Finetuning}} & & \multicolumn{2}{c}{\textbf{\textcolor{green}{\cmark} Finetuning}} \\ \cmidrule{2-3}\cmidrule{5-6}
 \textbf{\makecell{Method}} &  \textbf{GBT} & \textbf{MLP} && \textbf{GBT} & \textbf{MLP}\\
\midrule
\multicolumn{6}{c}{Lensless}\\
\midrule

\cite{dapueto2024transferring}&$70.32\pm0.029$&$71.93\pm0.030$&&$73.04\pm0.024$&$75.48\pm0.027$ \\

Our &$77.06\pm0.020$&$77.46\pm0.022$&&$93.55\pm0.019$&$94.62\pm0.017$ \\
\midrule
\multicolumn{6}{c}{WHOI15-2007}\\
\midrule

\cite{dapueto2024transferring}&$49.90\pm0.014$&$48.20\pm0.018$&&$50.98\pm0.016$&$49.29\pm0.020$ \\
Our&$47.92\pm0.015$&$51.96\pm0.023$&&$60.74\pm0.026$&$63.17\pm0.033$ \\

\midrule
\multicolumn{6}{c}{Vacuoles}\\
\midrule

\cite{dapueto2024transferring}&$64.03\pm0.041$&$59.89\pm0.053$&&$65.45\pm0.054$&$62.77\pm0.057$ \\
Our&$84.95\pm0.02$&$85.10\pm0.018$&&$90.45\pm0.019$&$89.97\pm0.019$ \\

\midrule
\multicolumn{6}{c}{Sipakmed}\\
\midrule

\cite{dapueto2024transferring}&$52.63\pm0.043$&$51.25\pm0.050$&&$55.10\pm0.041$&$55.69\pm0.038$ \\
Our&$61.75\pm0.019$&$63.33\pm0.014$&&$71.17\pm0.025$&$72.98\pm0.022$ \\

\bottomrule
\end{tabular} 
}
\label{tab:classification_all}

\end{table}

\subsection{Implementation details}

\noindent Since the images in all datasets are of different dimensions, we padded them to preserve the original aspect ratio and then resized to $224\times224$.

 We train Ada-GVAE on the synthetic Source Texture-dSprite dataset, using pairs of images that differ in $k=1$ factors of variation according to \cite{locatello2020weakly}, where this was shown to lead to higher disentanglement.
Following \cite{dittadi2020transfer,dapueto2024transferring}, we vary the parameter $\beta$ in $\{1, 2\}$.

We produce 20 Source models (10 random seeds $\times$ 2 values of $\beta$, latent dimension 10 for all) with the Adam optimizer and default parameters, batch size=64 and 400k steps. We use linear deterministic warm-up \cite{dittadi2020transfer, sonderby2016ladder} over the first 50k training steps. 
For the unsupervised finetuning on the microscopy dataset we used $\beta$-VAE, we finetuned the model for 20 epochs.

\begin{figure}[tb]
\centering
\resizebox{1.0\linewidth}{!}{
    \begin{subfigure}[t]{0.50\linewidth}
    \includegraphics[width=\linewidth]{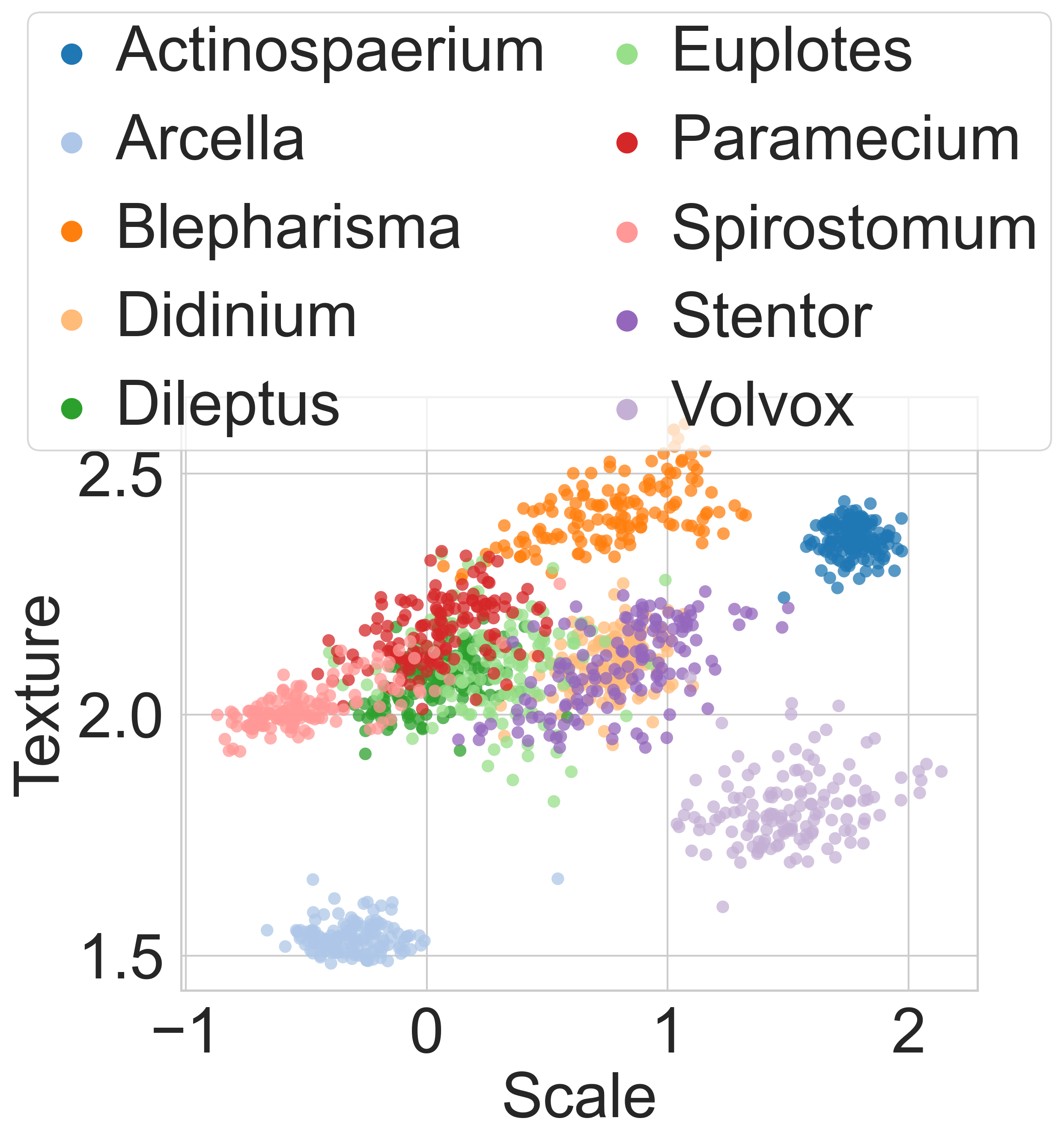}
    \caption{}
    \label{fig:scaletext}
    \end{subfigure}
    \hfill 
    \begin{subfigure}[t]{0.48\linewidth}
    \includegraphics[width=\linewidth]{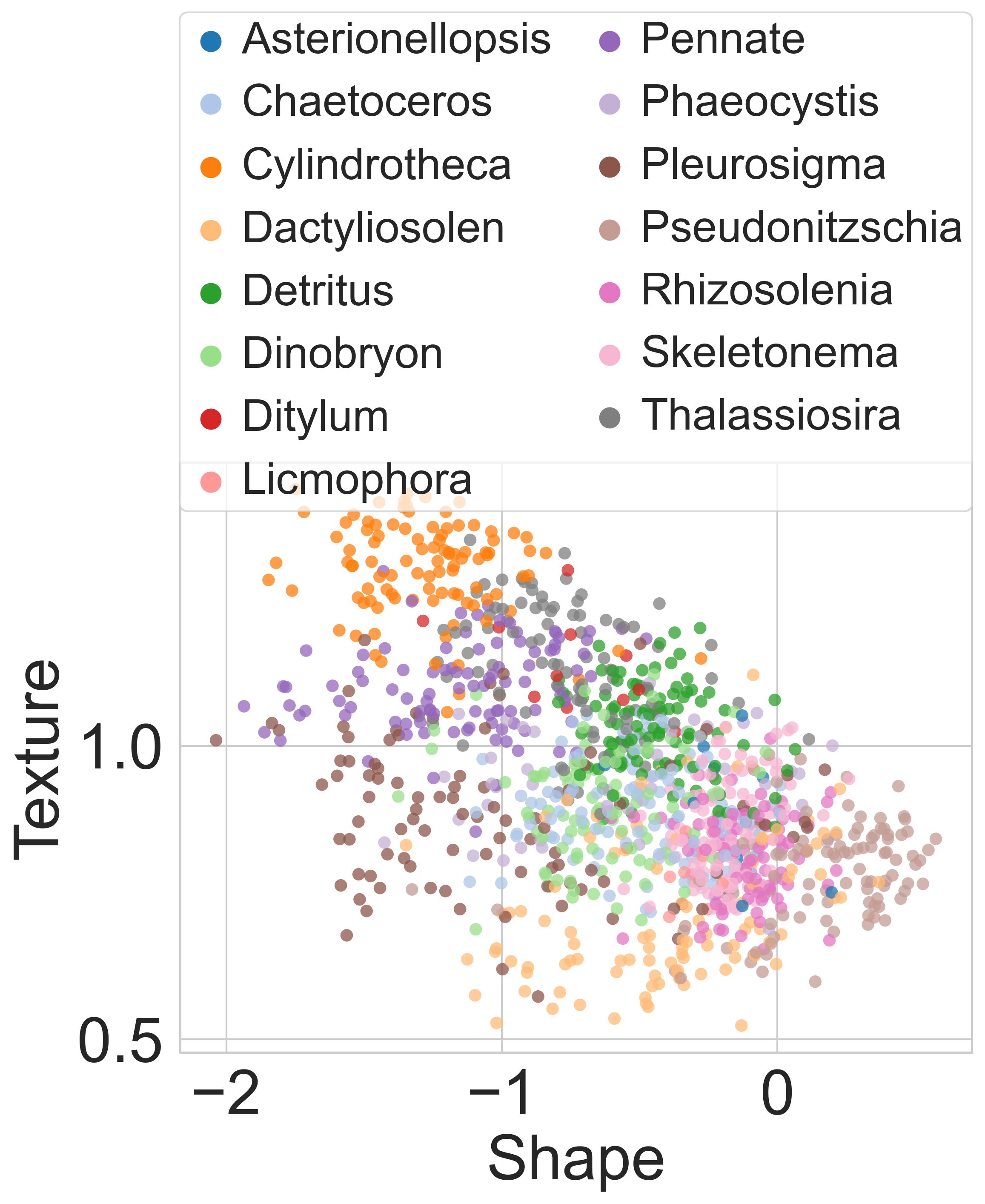}
    \caption{}
    \label{fig:shapetext}
    \end{subfigure}
}
    
\caption{Representation of Lensless (Fig. \ref{fig:scaletext}) and WHOI15 (Fig. \ref{fig:shapetext}) using the two most important features (finetuned models with input $\Phi$). }

\label{fig:lensless_tsne}

\end{figure}

\subsection{Evaluation protocol} 
We summarise the key elements of our experimental analysis, whose aim is to assess the potential of DRL of microscopy images, reported in the remainder of the section. 
{ We start from the disentangled representation $z$ (see Fig. \ref{fig:pipeline}) and remove the ``inactive'' dimensions with a standard deviation below a certain threshold{ (=0.05)}.

\noindent\textbf{Downstream task performances.} 
To assess the efficacy of our representation with respect to the microscopy classification downstream tasks,  we adopted two simple classifiers, a Gradient Boosted Trees (GBT) \cite{friedman2001greedy} and a Multilayer Perceptron (MLP) \cite{DBLP:journals/ijns/Lippmann94} with 2 hidden layers of size 256{, to better appreciate the influence of the representation on the performance in the downstream task. On the specific choices, we referred to \cite{dittadi2020transfer}.} 
{ Classification is evaluated using accuracy (since all the Targets are class-balanced) reporting mean and standard deviation over the 20 models. We analyse the results without and with finetuning (marked with \textcolor{red}{\xmark} and \textcolor{green}{\cmark}, in the tables). }

\noindent\textbf{Disentanglement.} We measure the quality of the disentanglement of the learnt models after the transfer. Since the real-world Target Datasets do not have any labels of the FoVs, we evaluate the disentanglement on Texture-dSprites (Source dataset) before and after the finetuning. This allows us to evaluate the "persistence" of the disentanglement after the finetuning.
Target datasets do not contain all the possible combinations of their FoVs and the latter do not exhibit independence, strictly required to learn disentangled representation, but with our transfer method, we expect the disentanglement to be preserved.{ We report the scores of the most used metrics in the literature -- MIG and DCI -- and the very recent OMES. They will be discussed in the next section.}

\noindent\textbf{Interpretability.} We report the normalized {\em feature importance} of the GBT models using the Gini importance \cite{nembrini2018revival}.{ We also provide qualitative analysis to inspect the values of the latent representations and their connection with the FoVs in the Target dataset. 

\subsection{Classification and interpretability assessment}

\textbf{Lensless:} 
Table \ref{tab:classification_all} shows the results obtained on the downstream task of plankton classification, on the disentangled representation learnt from Texture-dSprites. 
We compare different inputs on Ada-GVAE: { {(1)} the original RGB images as proposed in \cite{dapueto2024transferring}; {(2)} our approach with the $768$-dimensional deep features $\Phi$ produced by DINO\cite{Caron_2021_ICCV}}. The latter produces significantly higher performances.

The table also compares performances before and after finetuning. Considering the huge gap between Source and Target datasets, finetuning appears to be essential, although its benefit is more evident when using $\Phi$. The rich pretrained features provide very high results already without finetuning.\\
Figures \ref{fig:importance_NFT} and \ref{fig:importance_FT}  show the \emph{feature importance} before and after finetuning.
We can observe that after finetuning, it may change, nicely adapting to the specificity of the dataset, {where \texttt{Scale} and \texttt{Texture} are more relevant. }
\begin{figure}[tb]
\centering
\resizebox{1.0\linewidth}{!}{
    \begin{subfigure}[t]{0.49\linewidth}
    \includegraphics[width=\linewidth]{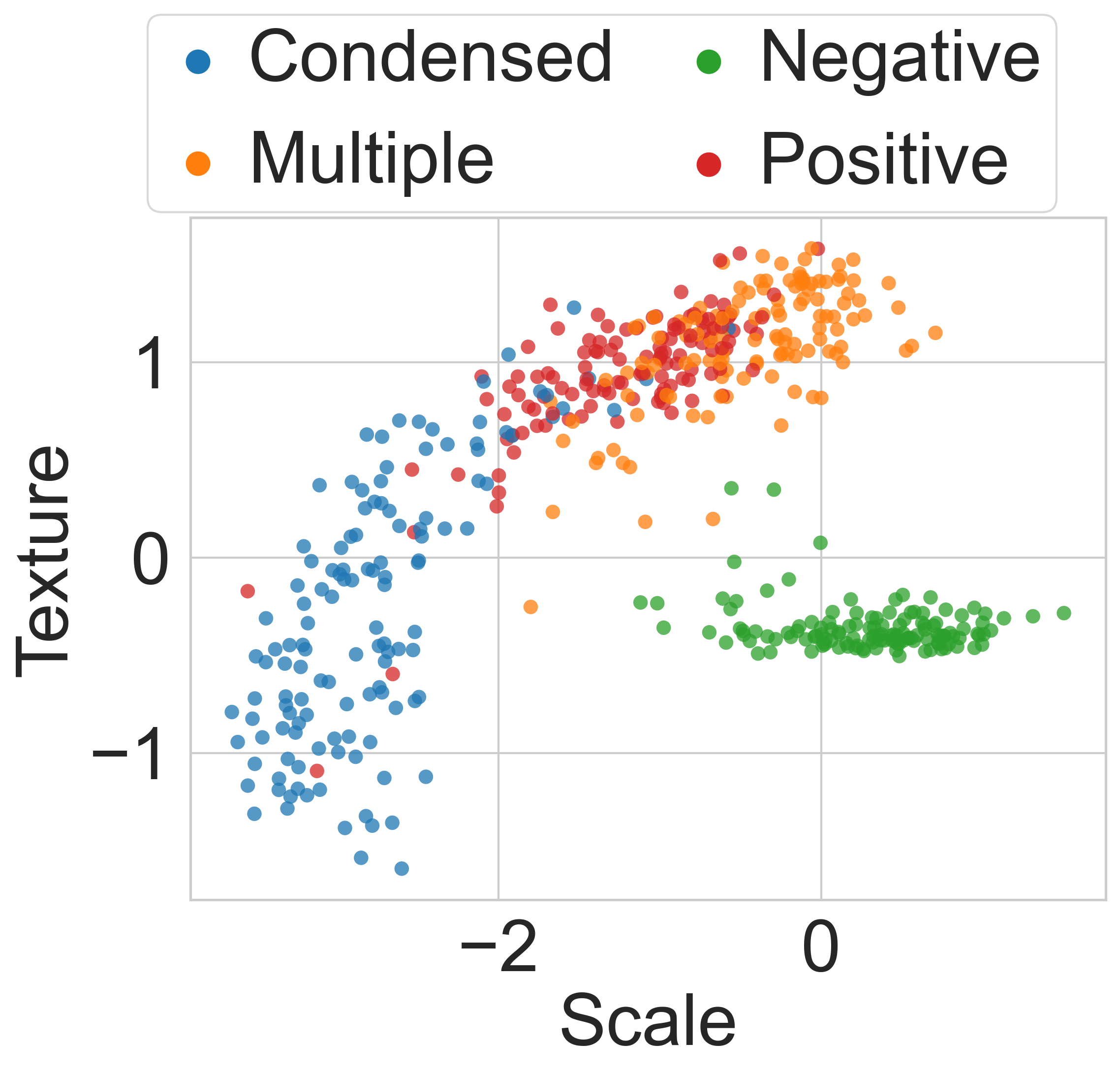}
    \caption{}
    \label{fig:scatter_vacuoles}
    \end{subfigure}
    \hfill 
    \begin{subfigure}[t]{0.49\linewidth}
    \includegraphics[width=\linewidth]{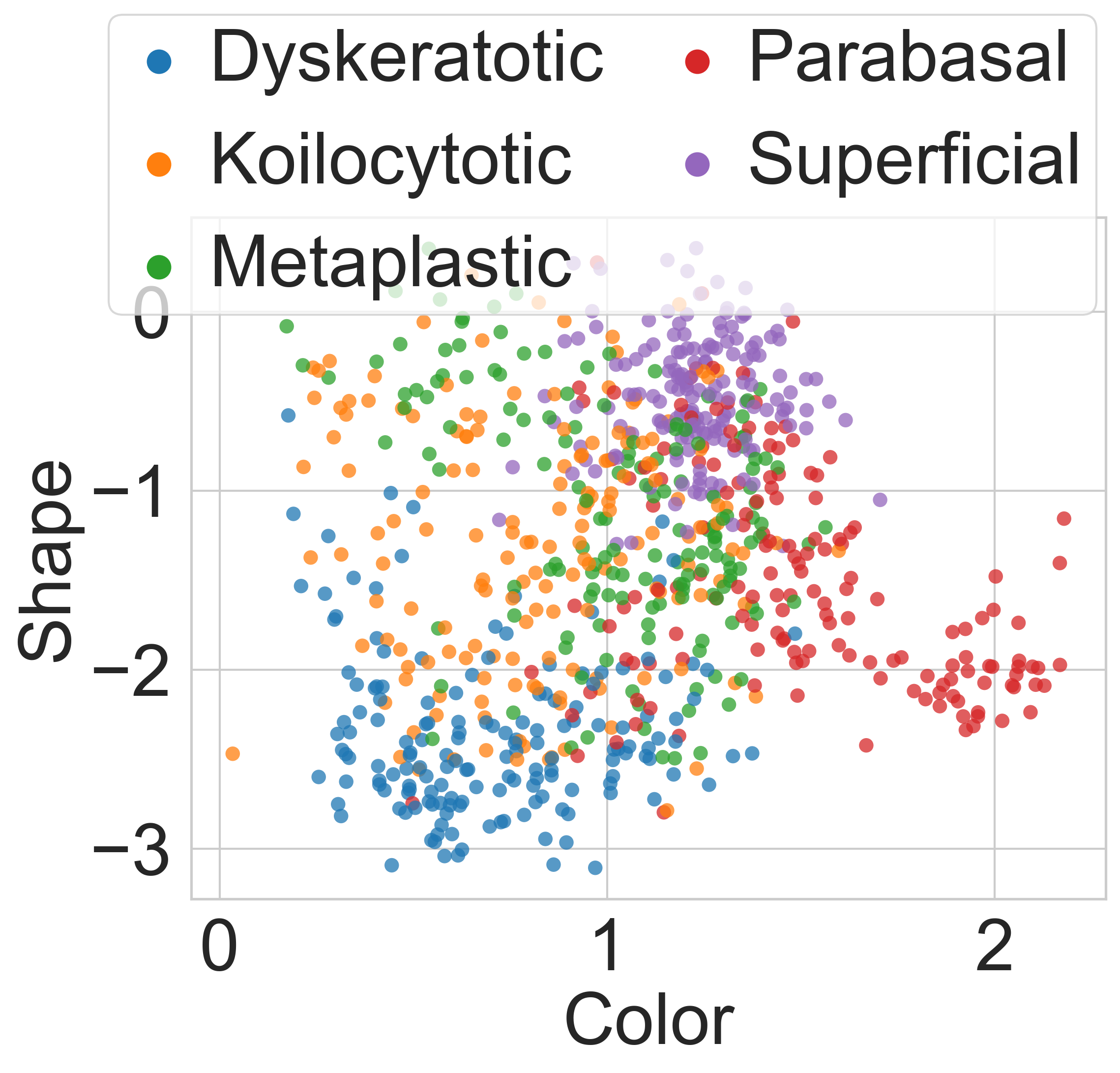}
    \caption{}
    \label{fig:scatter_sipakmed}
    \end{subfigure}
}
    
\caption{Representation of Vacuoles (Fig. \ref{fig:scatter_vacuoles}) and Sipakmed (Fig. \ref{fig:scatter_sipakmed}) using the two most important features (finetuned models with input $\Phi$).}

\label{fig:vacuoles_scatter}

\end{figure}

{For this reason, in Fig. \ref{fig:scaletext} we report a scatter plot of these features in our representation.}
{We observe that some classes (i.e. \textit{Actinospaerium}, \textit{Arcella}, \textit{Blepharisma}, \textit{Volvox}) need just these 2 features to be clearly separable from the others, while for the others (such as \textit{Didinium} and \textit{Stentor}) the 2 features are less distinctive.} \\
\noindent To provide an insight into the interpretability of the disentangled representation, we assess in Fig. \ref{fig:correlation} the Pearson correlation of the disentangled representation with a handcrafted one obtained by exploiting the mask ground-truth provided with the dataset to derive a set of features. Specifically, we computed a \texttt{scale} feature (as the area of the mask),\texttt{color} features (color average in the foreground) and a \texttt{shape} feature (solidity, as the ratio of the mask area and its convex hull) \cite{pastore2020annotation}.

Then, we computed the correlation between handcrafted and learned (disentangled) features. For the latter, we identified the latent dimension in the disentangled representation better encoding scale, color and shape (according to the annotated source dataset).

Handcrafted and latent scale features (Fig. \ref{fig:scalescale}) have a high correlation (-0.91) 
while handcrafted vs learnt color (we used the average red as an example in Fig. \ref{fig:colorred}) exhibit a milder correlation (-0.72). 
Solidity features have a smaller correlation with the dimension encoding the \texttt{Shape} factor (Fig. \ref{fig:shapesolidity}).{ This might also suggest that the complexity of the shape concepts can be hardly encoded in a single (handcrafted or learnt) value.}

\begin{figure}[tb]
\centering
    \begin{subfigure}[t]{0.29\linewidth}
    \includegraphics[width=\linewidth]{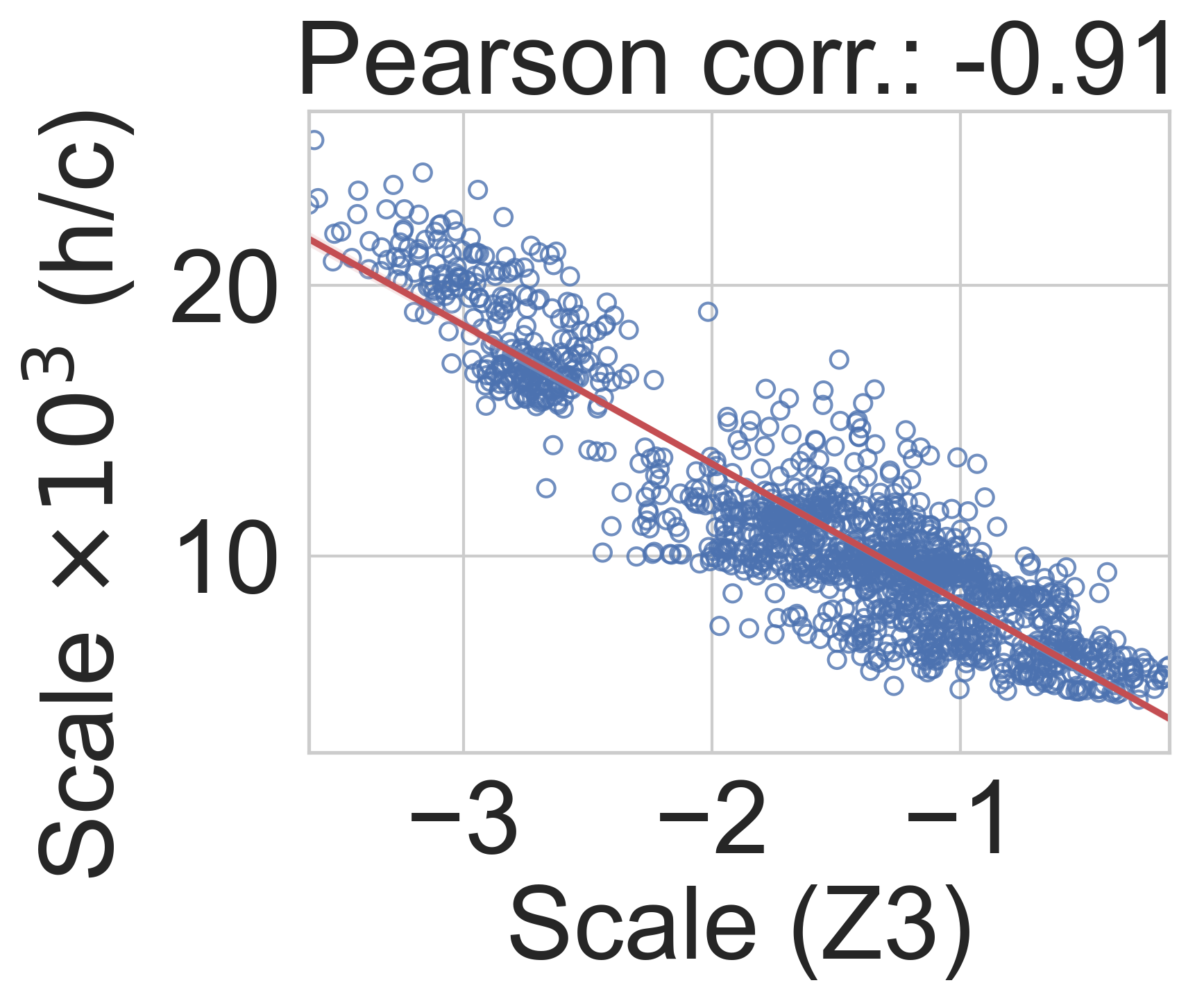}
    \caption{}
    \label{fig:scalescale}
    \end{subfigure}
    \hfill 
    \begin{subfigure}[t]{0.3\linewidth}
    \includegraphics[width=\linewidth]{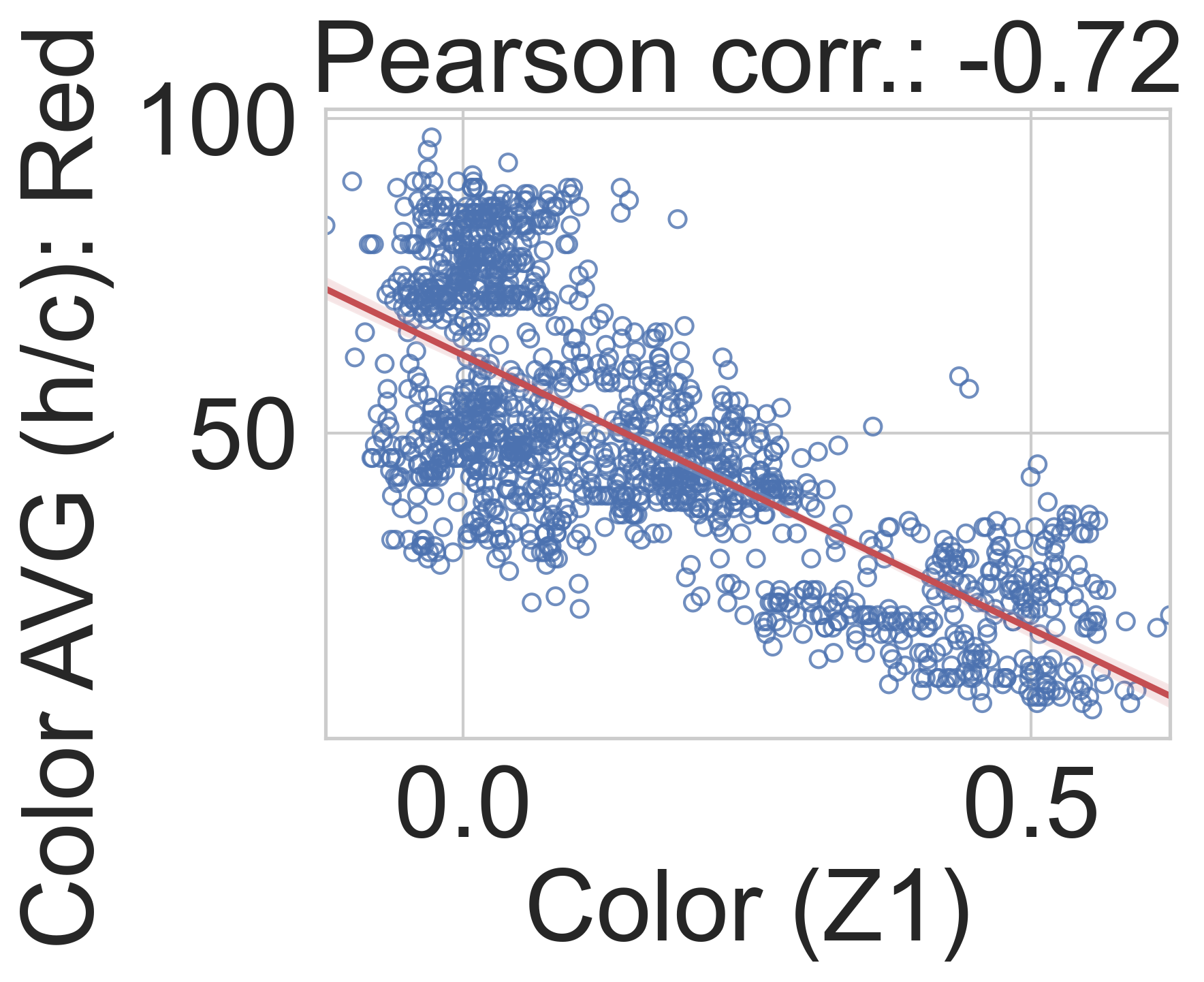}
    \caption{}
    \label{fig:colorred}
    \end{subfigure}
    \hfill
    \begin{subfigure}[t]{0.32\linewidth}
    \includegraphics[width=\linewidth]{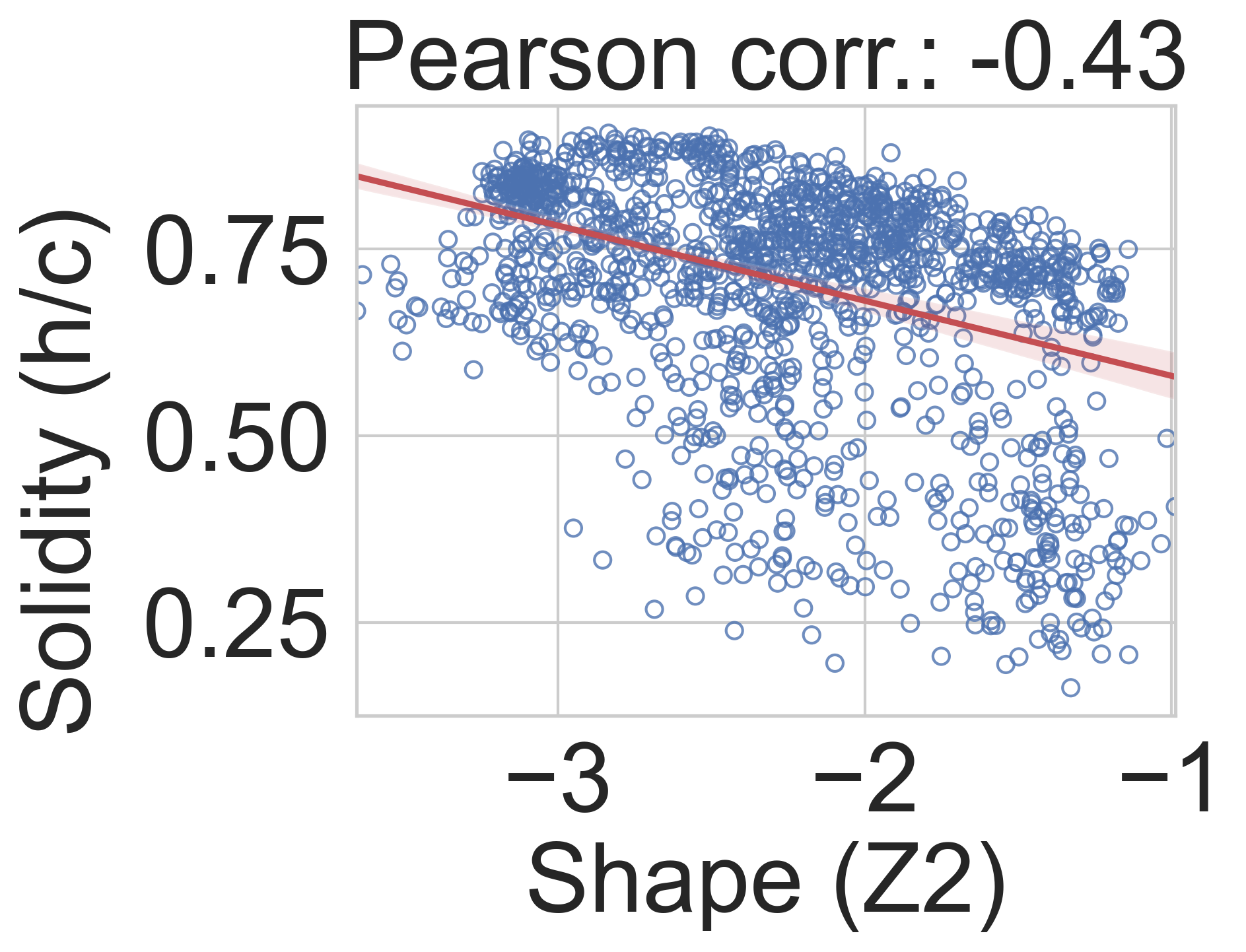}
    \caption{}
    \label{fig:shapesolidity}
    \end{subfigure}
    \caption{Correlation between our representation and handcrafted {(h/c)} \textit{scale} (Fig. \ref{fig:scalescale}), \textit{red channel} (Fig. \ref{fig:colorred}), and \textit{solidity} (Fig. \ref{fig:shapesolidity}), computed from the Lensless samples, as in \cite{pastore2020annotation}.}
\label{fig:correlation}

\end{figure}

\textbf{WHOI15-2007:}  Table \ref{tab:classification_all} reports the results obtained on this second, more complex, plankton dataset.

We notice a significant increase in performances, as we change the input from RGB to $\Phi$ in the presence of finetuning; without finetuning, the improvement is marginal or missing
 because of the high intra-class and extra-class variability of the dataset that our Source dataset cannot easily capture.

 Figure \ref{fig:lensless_feature_importance} shows the features importance before (Fig. \ref{fig:importance_NFT}) and after finetuning (Fig. \ref{fig:importance_FT}). From the latter we observe that the \texttt{Color} factor is the least important considering the dataset is nearly monochromatic, while the \texttt{Texture} and \texttt{Shape} are the most important ones.
\noindent Fig. \ref{fig:shapetext} shows the distribution of the first and second most important features  (\texttt{Texture} and \texttt{Shape}), where again we may appreciate how data are nicely clustered even for a higher number of classes.

\begin{table*}[tb]
\caption{Disentanglement score (\%) of Source and Finetuned models trained with our and {\cite{dapueto2024transferring}} methods.}
\centering
\resizebox{\linewidth}{!}{
\begin{tabular}[t]{cccccccccc}
\toprule
 && \multicolumn{2}{c}{\textbf{OMES ($\uparrow$)}} & & \multicolumn{2}{c}{\textbf{DCI ($\uparrow$)}} & & \multicolumn{2}{c}{\textbf{MIG ($\uparrow$)}} \\ \cmidrule{3-4} \cmidrule{6-7} \cmidrule{9-10}
 \textbf{\makecell{Dataset}} & \textbf{\makecell{Finetune}} & \textbf{\cite{dapueto2024transferring}} & \textbf{Our} && \textbf{\cite{dapueto2024transferring}} & \textbf{Our} && \textbf{\cite{dapueto2024transferring}} & \textbf{Our}\\

\midrule

Texture-dSprites&\textcolor{red}{\xmark}&$55.05\pm0.047$&$54.97\pm0.015$&&$62.09\pm0.028$&$43.75\pm0.029$ &&$42.13\pm0.033$&$40.18\pm0.026$\\
\midrule

Lensless &\textcolor{green}{\cmark}&$39.14\pm0.048$&$52.52\pm0.018$&&$33.21\pm0.051$&$37.47\pm0.026$  &&$20.31\pm0.039$&$35.05\pm0.024$\\

WHOI15-2007&\textcolor{green}{\cmark}&$46.98\pm0.052$&$51.06\pm0.020$&&$49.29\pm0.057$&$35.23\pm0.033$ &&$30.81\pm0.029$&$33.12\pm0.030$ \\

Vacuoles&\textcolor{green}{\cmark}&$48.54\pm0.062$&$54.08\pm0.015$&&$51.21\pm0.069$&$41.15\pm0.031$ &&$32.34\pm0.050$&$38.34\pm0.025$ \\

Sipakmed&\textcolor{green}{\cmark}&$31.16\pm0.043$&$51.00\pm0.053$&&$21.03\pm0.083$&$33.34\pm0.032$ &&$12.49\pm0.047$&$31.32\pm0.027$ \\

\bottomrule
\end{tabular} 
}

\label{tab:disentanglement}

\end{table*}

\textbf{Vacuoles:} Table \ref{tab:classification_all} reports the results on the Vacuoles dataset. Similar to the above analysis,
we observe a benefit of finetuning and an improvement when deep features $\Phi$ are used.
Figure \ref{fig:lensless_feature_importance} shows the feature importance before (Fig. \ref{fig:importance_NFT}) and after finetuning (Fig. \ref{fig:importance_FT}). We observe that the FoV \texttt{Color} is the least important feature, suggesting that color (that codifies the vacuole's depth information) is the least discriminative feature for classifying the target morphotypes. Figure \ref{fig:scatter_vacuoles}  shows the representation of \texttt{Texture} and \texttt{Scale}, we can observe that the \textit{negative} class is aligned to the scale axis meaning the samples have the same texture but different scale, which is in line with a visual observation of the negative class in the dataset (see Fig. \ref{fig:vacuoles_samples}).

\textbf{Sipakmed:} Table \ref{tab:classification_all} shows the results obtained on the downstream task of cell classification. The analysis we can make on the results is coherent with the observations made for the previous datasets. In particular, once again we observe that the models trained on the deep feature $\Phi$ lead to a greater improvement when finetuned compared to the RGB-based one. 

Comparing our results with the ones of the original work \cite{plissiti2018sipakmed}obtained with handcrafted features and the MLP classifier (78.92\% of accuracy), we observe that we achieved slightly lower performances in terms of accuracy. The original work used handcrafted features describing intensity, texture and shape calculated  for both the region of the nucleus and the cytoplasm of each cell. Our learned factors take into consideration the entire image and cannot tell apart the specific information of each part of the cell. In order to improve the performances and the representation of this specific dataset, an ad-hoc Source dataset to take into account the FoVs of the separated parts of the cell would be useful. As confirmation of this, Fig. \ref{fig:importance_FT}  shows the features' importance after the finetuning being very similar, meaning that all the features have the same importance (except for \texttt{Shape}), suggesting that ad-hoc FoVs are required to provide a more in-depth disentanglement.
Figure \ref{fig:scatter_sipakmed} shows the representation of \texttt{Shape} and \texttt{Color}: while we observe that the \textit{Koilocytotic} class is quite scattered and partially overlapping with most classes, \textit{Dyskeratotic} and \textit{Superficial} are separated. However, they are all quite scattered, suggesting again that this dataset may need ad-hoc FoVs.

\underline{Discussion:}} 
 With an experimental analysis performed on four microscopy datasets with variable characteristics, we have shown that transferring a disentangled representation learned from a synthetic Source dataset to a real Target dataset is possible. We have also observed that the quality of the disentanglement and its meaningfulness for a downstream classification task may change depending on the input (RGB or deep features) and the adoption of a step of finetuning of the model to be transferred. { Moreover, we compared the disentangled models obtained from raw RGB images and the deep feature $\Phi$ extracted from a pretrained network (Vit16b pretrain with DINO)}. We could observe that the latter allows for a more robust transfer, further improving the performances of a disentangled representation that is also human-interpretable.

{
\subsection{Disentanglement evaluation experiments}

}
\label{sec:dis_discussion}

{Table \ref{tab:disentanglement} reports the disentanglement score for the metric OMES,  that measures both \textit{Compactness} and \textit{Modularity}.  The score without the finetuning is obtained from the original Source model trained with weak supervision, thanks to which a high level of disentanglement is obtained.  The scores referring to the Target datasets are computed by extracting the representation of Texture-dSprites using the different finetuned models since it is not possible to do the same directly on the Target for the lack of annotation.

Overall, we observe that the models trained with the deep features $\Phi$ (our) preserve the same level of disentanglement of the Source models independently on the Target dataset. On the other hand, the models trained with the images\cite{dapueto2024transferring} do not preserve the disentanglement of the Source model, and the level of degradation also depends on the Target dataset.
 {The analysis for the other metrics, MIG and DCI measuring the two properties separately (see Table \ref{tab:disentanglement}), is analogous}.\\
{Moreover, the level of disentanglement of the original DINO features (OMES=26\%, MIG=3\%) is lower than the one of our learnt latent representation (OMES=54.97\%, MIG=40.18\%), and hence our methodology enhances disentanglement even when applied to a rich initial representation. We did not include the metric DCI in this comparison since, as found in \cite{cao2022empirical}, DCI does not allow a fair comparison between representations of different latent dimensions.}\\
 This suggests that \textbf{transferring from deep features extracted from pretrained models is more robust and preserves the disentanglement} also across dataset of very different domains and very different from the Source dataset.
\begin{figure*}
  \centering
  \begin{minipage}[tb]{0.65\textwidth}
    \centering
    \captionof{table}{Explicitness of the FoV of Tetxure-dSprites of different pretrained networks. In \textbf{bold} the best performing model.}
    \label{tab:backbones}
   \resizebox{\linewidth}{!}{
\begin{tabular}{lcccccccc}
\toprule
          &  \multicolumn{7}{c}{\textbf{Tetxure-dSprites FoVs (\%)}}\\\cmidrule{2-8}
 \textbf{Pretraining}& \textbf{Texture}& \textbf{Color}& \textbf{Shape}& \textbf{Scale}& \textbf{Orient.}& \textbf{PosX}&\textbf{PosY} & \textbf{All}\\
 \midrule

  \textit{Random}&  20.00 & 14.28 & 33.33 & 16.66 & 2.5 & 3.12 & 3.12 & 13.28\\
  ResNet152 &  92.50 & 96.96 & 56.93 & 53.03 & 6.17 & 16.49 & 17.52 & 48.51 \\
 DenseNet201 &   96.11 & 96.78 & 77.57 & 47.67 & 8.52 & 19.09 & 19.89 & 52.23 \\
VGG19 & 97.68  & 86.24 &  87.99 & 69.87 & 19.50 & 27.49 & \textbf{38.02}  & 60.97 \\\midrule
  Swing4-Large & 32.80 & 21.74 & 33.39 & 21.94 & 2.50 & 8.63 & 17.21 & 19.74\\
  Vit16-Large &  99.44 & 99.98 & 93.56 & 79.56 & 20.85 & 19.35 & 17.83 & 61.51 \\
  Vit16-Base + DINO &  \textbf{99.74} & \textbf{100.00} & \textbf{95.36} & \textbf{88.78} & \textbf{34.00} & \textbf{31.9} & 34.74 & \textbf{69.21}\\

\bottomrule
\end{tabular}
}
  \end{minipage}\hfill
  \begin{minipage}[tb]{0.35\textwidth}
    \centering
    \captionof{table}{ Accuracy ($\%$) and SD over 20 classifiers { trained directly on $\Phi$, removing the VAE.}}
    \label{tab:ablation_disentanglement}

    \resizebox{\linewidth}{!}{
\begin{tabular}{ccc}
\toprule
\textbf{Dataset} &  \textbf{GBT} & \textbf{MLP} \\
\midrule
 Lensless &  $99.13\pm0.002$ & $99.49\pm1.11$ \\
 WHOI15-2007 &  $84.70\pm0.01$ & $97.16\pm0.01$ \\

Vacuoles &  $94.52\pm0.01$ & $97.46\pm0.04$ \\
Sipakmed &  $92.09\pm0.005$ & $94.95\pm0.01$ \\

\bottomrule
\end{tabular} 
}
  \end{minipage}
\end{figure*}
{
\subsection{Ablation study} \label{ablation}
 On the specific choice of the VIT pretrained with DINO self-supervised approach, in Table \ref{tab:backbones} we report the \textit{Explicitness} of pretrained models on ImageNet1K, on the FoVs of Texture-dSprites. It can be observed that our choice outperforms the other models.\\
In Table \ref{tab:ablation_disentanglement}, we report an ablation study in which we removed the disentanglement, directly employing the deep features $\Phi$ for the downstream classification tasks. In this way, we can quantify how much we lose in terms of accuracy when advancing interpretability.{We can observe that for WHOI15, the disentanglement degrades the classification performances (see Table \ref{tab:classification_all}). WHOI15 is a dataset containing multiple cells per image, making the data more complex and for this reason it may need further FoVs to be represented and disentangled.}
{Moreover, it is worth noting that our disentangled features are of much smaller size, 10, and they require fine-tuning to reach comparable results with non-disentangled features.}}  This shows that the disentanglement of the model, which provides a level of interpretability, comes with a price.

\section{Preliminary assessment on open set classification}
To provide a specific application example, showing the potential of interpretability {in the field of microscopy images}, we consider anomaly detection as a way to use plankton as a biosensor \cite{pastore2019establishing, ciranni2024anomaly}. Anomalies can either correspond to novel classes or {\em plankton organisms reacting to environmental perturbations} \cite{ciranni2024anomaly}. We aim to assess whether DRL can provide further information on samples detected as anomalous, allowing us to distinguish between the two described scenarios.  
As a study case, we remove the \textit{Arcella Vulgaris} samples from the Lensless training set and use the remaining 9 classes for finetuning and species classification.

We then feed \textit{Arcella} images to the classifier, which predicts them as \textit{Euplotes Eurystomus}, with high confidence. 
 We compute the mean distance for each dimension of the disentangled representation from the \textit{Arcella} samples to the centroid of \textit{Euplotes}, finding that \texttt{Shape} (1.42)  and \texttt{Texture} (0.95) are the most further dimensions and \texttt{Color} (0.18) and \texttt{Scale} (0.27) are the closest. Fig. \ref{fig:anomaly} shows how \textit{Arcella} (red), \textit{Eupotes} (blue), and Others (grey) samples are encoded and despite \textit{Arcella} being classified as \textit{Euplotes}, we can appreciate the distance between the two classes in the \texttt{Texture}-\texttt{Shape} space (\ref{fig:textshapeOS}), while the classes almost overlap in the \texttt{Color}-\texttt{Scale} space (\ref{fig:colorscaleOS}). This provides insights on the actual difference between the FoVs of our test samples and the training class they are assigned to by the classifier, which could be used to identify unseen or anomalous classes in the described application scenario. 
\begin{figure}[tb]
\centering
\resizebox{1.0\linewidth}{!}{
    \begin{subfigure}[t]{0.50\linewidth}
    \includegraphics[width=\linewidth]{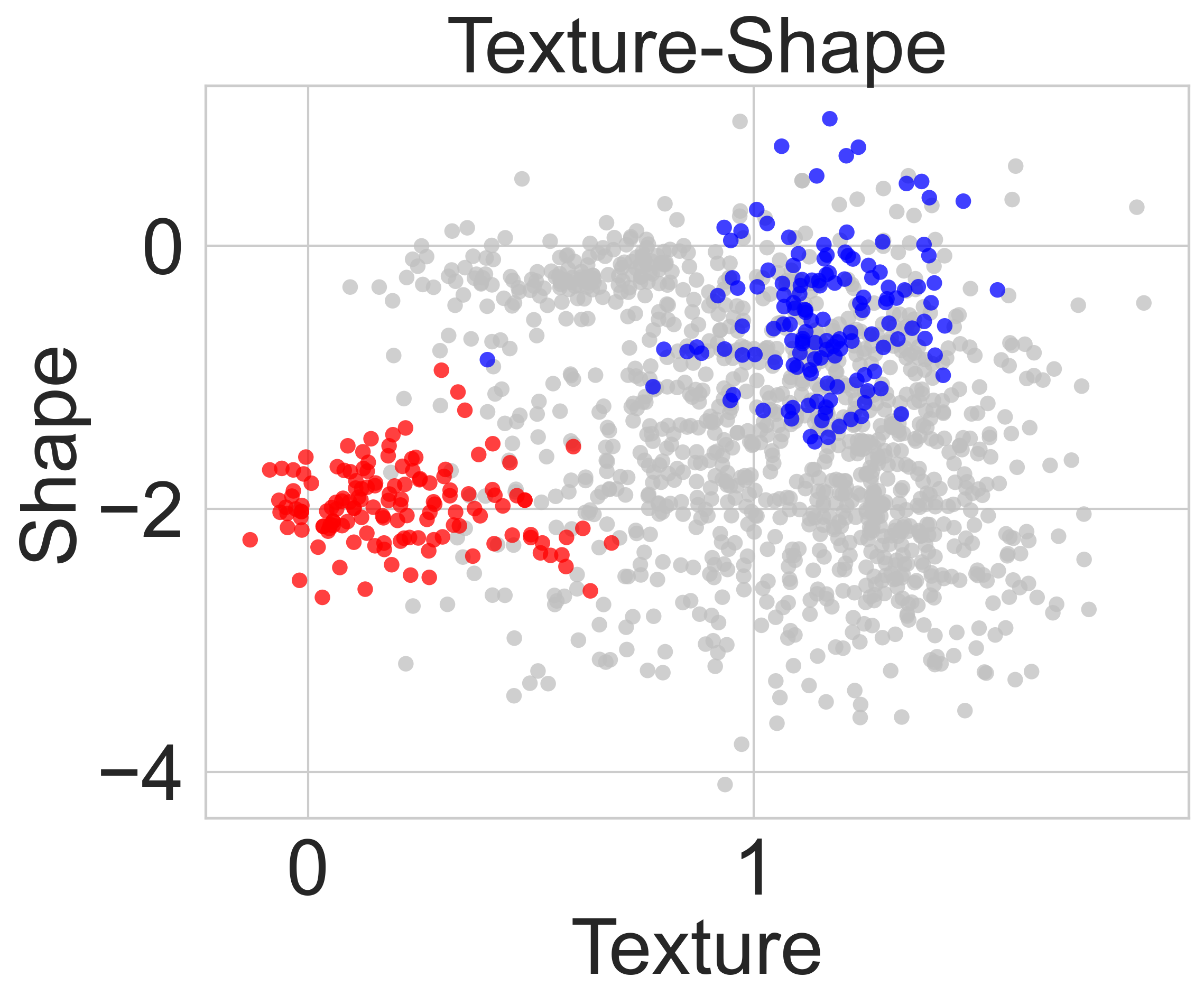}
    \caption{}
    \label{fig:textshapeOS}
    \end{subfigure}
    \hfill 
    \begin{subfigure}[t]{0.48\linewidth}
    \includegraphics[width=\linewidth]{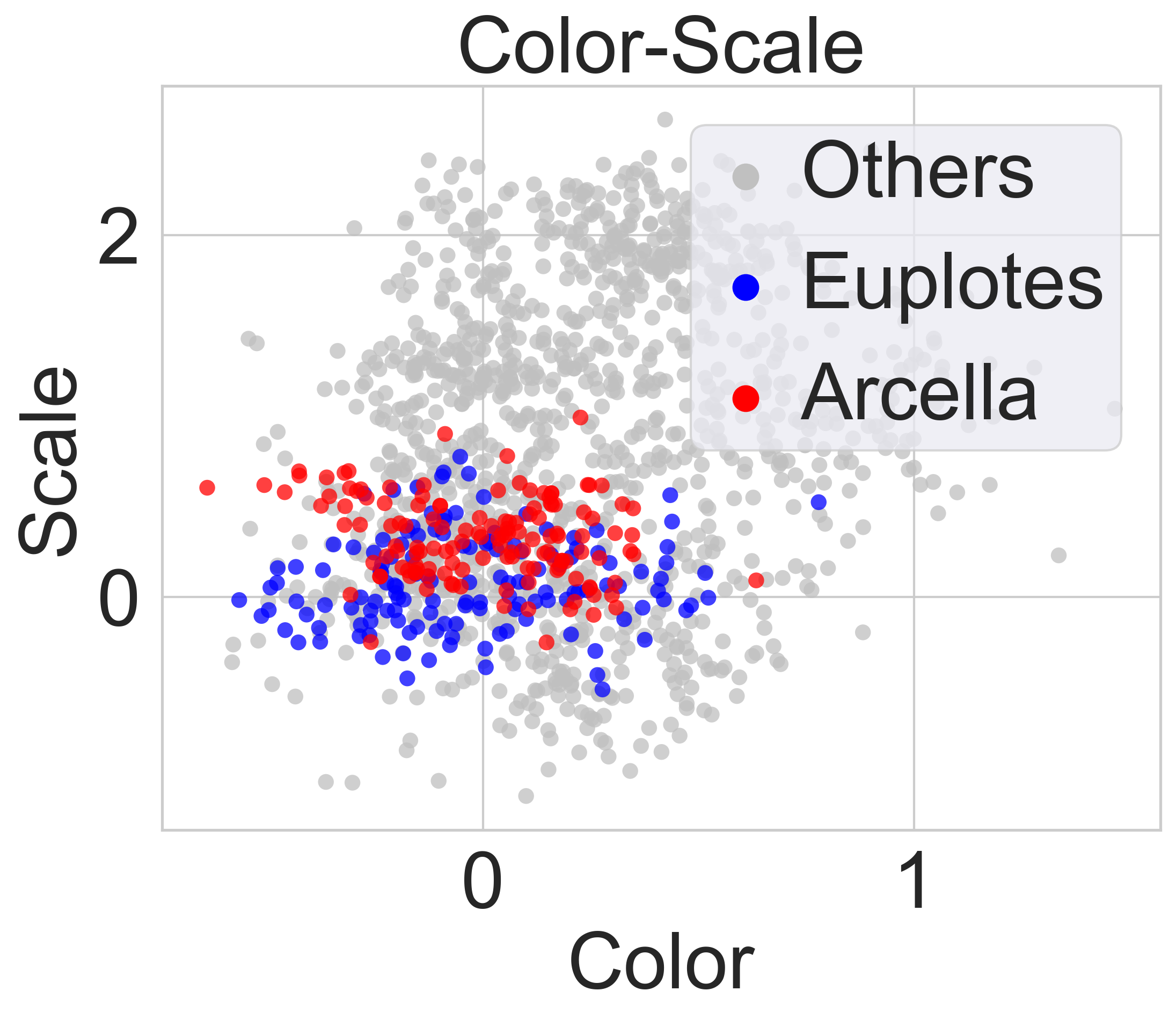}
    \caption{}
    \label{fig:colorscaleOS}
    \end{subfigure}
}
\caption{The representation in the \texttt{Texture}-\texttt{Shape} (Fig. \ref{fig:textshapeOS}) 
 and \texttt{Color}-\texttt{Scale} spaces (Fig. \ref{fig:colorscaleOS}). \emph{Arcella} and \emph{Euplotes} are separated  or overlapped  depending on the features.}
\label{fig:anomaly}
\end{figure}

\section{Conclusion}\label{conclusion}

In this work, we presented a study on disentangled representation learning for microscopy images,  disentangling morphological factors such as texture, color, shape, and scale. 
Our results on four different microscopy benchmarks suggest that the learned disentangled representations provide a good trade-off between classification accuracy and interpretability, with a finetuning protocol being particularly beneficial when deep pretrained features are used as input data.\\
{\underline{Limitations and future work.}}
For the time being our analysis only includes only VAE-based methods, one could carry out an analogous study with more complex and powerful methods, such as Diffusion Models. 
We considered a general-purpose Source dataset which may not perfectly fit the FoVs of the Target dataset, for example as observed for Sipakmed. Future work will also consider the possibility of generating a synthetic FoV annotated dataset more specific to the purpose.

{ {\bf Acknowledgements.}  We acknowledge the financial support from PNRR MUR Project PE0000013 "Future Artificial Intelligence Research (FAIR)", funded by the European Union – NextGenerationEU, CUP J33C24000430007}

\bibliographystyle{ieeetr}
\bibliography{reference}
\end{document}